\documentclass{article}

\usepackage[numbers]{natbib}

\usepackage[position, final]{neurips_2026}

\usepackage[utf8]{inputenc} 
\usepackage[T1]{fontenc}    
\usepackage{hyperref}       
\usepackage{url}            
\usepackage{booktabs}       
\usepackage{amsfonts}       
\usepackage{nicefrac}       
\usepackage{microtype}      
\usepackage{xcolor}         

\usepackage{graphicx}
\usepackage{wrapfig}
\usepackage{subcaption}
\usepackage{enumitem}
\usepackage[breakable]{tcolorbox}

\usepackage[table]{xcolor}

\usepackage{amssymb}
\usepackage{amsmath,amsfonts}
\usepackage{amsopn}
\usepackage{bm} %
\usepackage{multirow}
\usepackage{tabularx}

\newcommand{\ProbOpr}[1]{\mathbb{#1}}

\newcommand{\expect}[2]{%
\ifthenelse{\equal{#2}{}}{\ProbOpr{E}_{#1}}
{\ifthenelse{\equal{#1}{}}{\ProbOpr{E}\left[#2\right]}{\ProbOpr{E}_{#1}\left[#2\right]}}} %
\newcommand{\var}[2]{%
\ifthenelse{\equal{#2}{}}{\ProbOpr{VAR}_{#1}}
{\ifthenelse{\equal{#1}{}}{\ProbOpr{VAR}\left[#2\right]}{\ProbOpr{VAR}_{#1}\left[#2\right]}}} %

\newcommand{\eat}[1]{}

\newcommand\mypara[1]{\vspace{.1cm}\noindent\textbf{#1\xspace}}

\definecolor{egogreen}{RGB}{214,235,197}
\definecolor{refred}{RGB}{251,210,204}
\definecolor{LightCyan}{rgb}{0.88,1,1}
\definecolor{LightRed}{rgb}{1,0.94,0.94}
\definecolor{LightBlue}{rgb}{0.94,0.97,1}
\definecolor{scarlet}{RGB}{147,0,0}
\definecolor{citeblue}{RGB}{238,26,28}
\definecolor{Blue}{RGB}{51,10,154}

\definecolor{citecolor}{RGB}{50, 63, 138}
\definecolor{linkcolor}{RGB}{187,18,26}

\RequirePackage{xspace}
\makeatletter
\DeclareRobustCommand\onedot{\futurelet\@let@token\@onedot}
\def\@onedot{\ifx\@let@token.\else.\null\fi\xspace}

\def\eg{\emph{e.g}\onedot}

\def\vs{\emph{vs}\onedot}

\makeatother

\usepackage{cleveref}
\hypersetup{
    colorlinks=true,
    linkcolor=[RGB]{44,95,138},   
    citecolor=[RGB]{26,111,186},  
    urlcolor=[RGB]{26,111,186},
}

\newcommand\extrafootertext[1]{%
    \bgroup
    \renewcommand\thefootnote{\fnsymbol{footnote}}%
    \renewcommand\thempfootnote{\fnsymbol{mpfootnote}}%
    \footnotetext{#1}%
    \egroup
}

\title{Autonomous Driving Research Requires a Community-Driven Data Paradigm}

\author{%
  Jinsu Yoo$^{1\dagger}$ \quad Zanming Huang$^{1\dagger}$ \quad Katie Z Luo$^2$ \quad Zheda Mai$^{3}$ \quad Qiyuan Wu$^{4}$\\[0.3em]
  {\bf Bharath Hariharan$^{4}$ \quad Mark Campbell$^{4}$ \quad Wei-Lun Chao$^{1}$} \\[0.5em]
  {\small $^1$Boston University \quad $^2$Stanford University \quad $^3$The Ohio State University \quad $^4$Cornell University}
}

\begin{document}

\maketitle

\extrafootertext{
$^\dagger$Equal contribution. Surveyed dataset list: \url{https://jinsuyoo.info/community-driving-data}
}

\begin{abstract}
Autonomous driving has made remarkable progress, with recent AI advances enabling commercial deployments that are reshaping urban mobility. Yet the field remains far from its \emph{universal} social promise: autonomous systems that can operate robustly \emph{anywhere, anytime, for anyone}. We posit that this gap is not merely a modeling problem, but a problem of the prevailing \emph{data paradigm}. Current research relies heavily on a few benchmark datasets with limited spatial and scenario coverage, even though the community has collectively produced over 600 autonomous driving datasets across nearly 50 countries. However, this abundance has not translated into broad research impact: most datasets remain significantly underused due to fragmentation, limited visibility, incompatible protocols, and benchmark incentives that concentrate attention on a few dominant datasets. We therefore argue that \textbf{autonomous driving research requires a collaborative, community-driven data paradigm}. Such a paradigm would improve the discovery, reuse, integration, and evaluation of diverse datasets; make underexplored data easier and more rewarding to study; and lower the barrier for new contributors. We outline its key principles, illustrate an early realization, and call for collaboration across academia and industry to transform fragmented datasets into shared community infrastructure for anytime-anywhere autonomy. 
\end{abstract}
\section{Introduction}
\label{sec:introduction}

Autonomous driving carries a compelling social promise: improving road safety~\cite{kusano2025comparison}, expanding mobility for the elderly~\cite{faber2020will}, people with disabilities~\cite{kassens2021autonomous}, and non-drivers, and ultimately enabling transportation that is available \emph{anywhere, anytime}. Recent advances in AI have brought this vision closer, accelerating commercial deployments and driving rapid progress in end-to-end self-driving models~\cite{hu2023uniad, jiang2023vad, weng2024paradrive, liao2025diffusiondrive}.

Despite this exciting progress, the \emph{anywhere, anytime} vision remains far from realized. Today's industry predominantly follows a city-by-city expansion strategy~\cite{waymo2025expansion}: before service can launch in a new city, massive efforts in data collection, model adaptation, safety validation, and infrastructure investment must take place---into a world of over 12,000 diverse cities~\cite{undesa2025wup}. This strategy is understandable, as safety and user experience are non-negotiable priorities in real-world deployment. As of May 2026, Waymo---arguably the leading provider of commercial robotaxi services in North America---serves riders in only 11 US metro areas or regions~\cite{waymo2026service}. The gap between the universal promise of autonomous driving and the pace of scalable deployment remains stark.

This deployment gap gives the research community an important complementary role: pushing the field toward broader geographic and operational generalization. Unlike commercial deployment, offline, pre-deployment research is not constrained by the same immediate safety, regulatory, and profitability requirements, allowing researchers to investigate diverse and challenging conditions: rare long-tail scenarios, low-density regions, and geographies that are commercially unattractive yet socially important. Broader coverage in data collection and evaluation could lay the groundwork for eventual real-world deployment at scale. Yet this potential remains largely unrealized. A survey of the autonomous driving literature reveals that research has overwhelmingly concentrated on a small set of benchmark datasets---KITTI~\cite{geiger2012kitti}, nuScenes~\cite{caesar2020nuscenes}, and the Waymo Open Dataset~\cite{sun2020waymo}---collectively spanning only a handful of cities across the US, Germany, and Singapore. In effect, the research community has inherited much of the same narrow scope that limits commercial deployment, rather than systematically expanding beyond it.

At first glance, this concentration appears to be a consequence of limited data availability. Compared with image and text data commonly used in computer vision and natural language processing, autonomous driving data are far harder to obtain: they require sensor-equipped vehicles operating in the physical world, precise calibration and synchronization across modalities, and dense annotation. It may therefore seem reasonable that the research community is constrained by publicly available datasets and must rely on major players, primarily large industrial actors, to lead data collection and curation~\cite{geiger2012kitti, sun2020waymo, caesar2020nuscenes, houston2021lyft, geyer2020a2d2}.

\begin{figure}[t]
  \centering
  \begin{subfigure}[b]{0.3\linewidth}
    \centering
    \includegraphics[width=\linewidth]{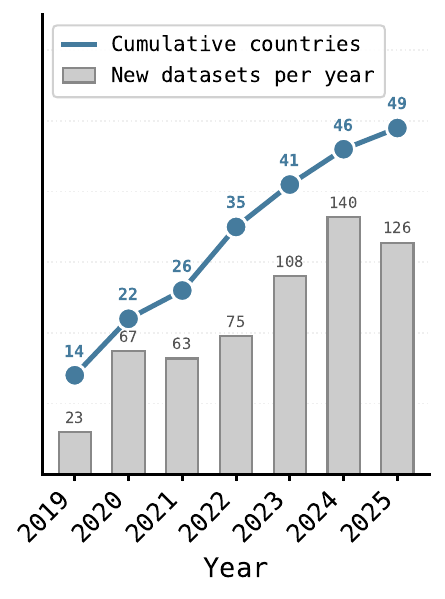}
    \vskip-10pt
    \caption{Cumulative country coverage.}
    \label{fig:countries_with_papers}
  \end{subfigure}
  \hspace{10pt}
  \begin{subfigure}[b]{0.57\linewidth}
    \centering
    \includegraphics[width=\linewidth]{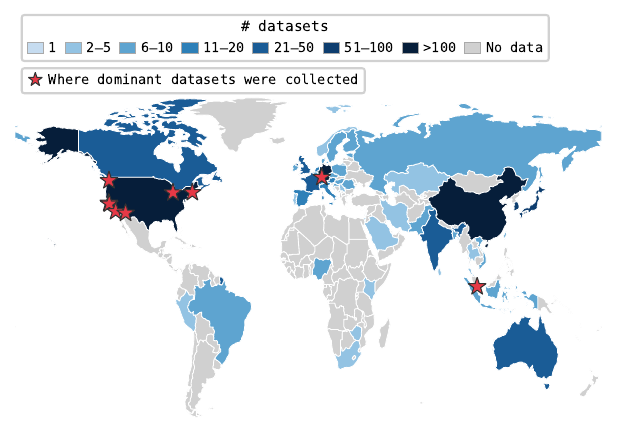}
    \vskip-10pt
    \caption{Geographic distribution of datasets across countries.}
    \label{fig:countries_world_map}
  \end{subfigure}
  \caption{\textbf{The community's driving data collection effort spans a broad and growing geographic footprint.} Coverage has expanded from a handful of countries in 2019 to nearly 50 by 2025, reaching all inhabited continents. The community has collectively built a diverse foundation of driving data. The $\star$ symbols mark the cities or regions where data for the most widely used benchmarks---KITTI~\cite{geiger2012kitti}, nuScenes~\cite{caesar2020nuscenes}, and the Waymo Open Dataset~\cite{sun2020waymo}---were collected.}
  \label{fig:geo_combined}
\end{figure}

Our targeted literature search, however, reveals a different picture. Since 2012, \emph{more than 600 real-world datasets} for autonomous driving have been published, with their collective spatial coverage steadily increasing from only a handful of countries to nearly fifty in recent years (\autoref{fig:geo_combined}). Importantly, many of these datasets were not collected by large industrial actors, but by the research community itself. Moreover, our preliminary evaluations on these community-contributed datasets reveal boundaries and limitations of state-of-the-art models trained primarily on industry-led datasets (see \autoref{sec:diverse-data-useful}). These findings suggest that the community has both the ability and the genuine interest to contribute to the data effort and expand its geographic and scenario diversity.

Yet these collective data efforts have not translated into broad research impact. 
Dataset usage follows a pronounced long-tail distribution: while a few benchmarks dominate model development and evaluation, most real-world driving datasets remain significantly underused (\autoref{fig:longtail}). 
This underuse, however, rarely reflects a lack of value---rather, it stems from practical barriers to adoption, including limited visibility, narrow modality coverage, small scale, difficult access, or uneven annotation quality. 
This is further exacerbated by the fact that algorithm developers are strongly incentivized to evaluate on widely recognized benchmarks, where comparisons are easier and publication expectations are clearer. 
Together, these barriers reveal a deeper structural problem: the community lacks a coordinated data ecosystem. Dataset contributors typically work in isolation, independently choosing sensor suites, collection sites, scenario definitions, annotation protocols, release formats, baseline tasks, and maintenance practices. As a result, datasets that collectively offer broad geographic and scenario coverage remain difficult to discover, compare, integrate, and reuse, leading to persistent data isolation. Model failures observed on one dataset rarely translate into systematic feedback for future data collection or benchmark design. Data expansion therefore remains inefficient: contributors lack clear signals about where new data would be most valuable, while newcomers face the burden of managing the entire pipeline from collection to release.

In this position paper, \textbf{we argue that the autonomous driving community requires a next-generation collaborative data paradigm that turns \textit{distributed data efforts} into \textit{shared community infrastructure}}. Such a paradigm should coordinate data collection, annotation, benchmarking, and maintenance across contributors while establishing shared protocols that make heterogeneous datasets easier to discover, compare, integrate, and reuse. Perhaps more importantly, it should also enable a feedback loop between model evaluation and data expansion, \textit{i.e.}, failures observed on existing datasets should inform targeted data collection, benchmark construction, and future community priorities. To ground this discussion, we present an early instantiation of what such a platform could look like, outline the principles it should embody, and identify the research opportunities it would enable. We also discuss practical challenges, including governance, licensing, privacy, and incentives for participation. Realizing this vision requires community-wide effort and agreement; no single group can do it alone. We therefore conclude with a call for collaboration across researchers, institutions, and industry toward a data paradigm that accelerates, rather than constrains, progress toward autonomous driving that works anywhere, for anyone.
\section{Background on Autonomous Driving Research}
\label{sec:background}

Autonomous driving research has spanned several decades~\cite{pomerleau1988alvinn}. Early milestones such as the DARPA Urban Challenge in 2007 and the launch of Google's self-driving car project in 2009 helped establish the feasibility of autonomous driving beyond controlled demonstrations. In the learning-based era that followed, much of the research community's progress was organized around component-level tasks, including perception, localization, tracking, prediction, and planning~\cite{chen2016monocular, chang2018pyramid, dosovitskiy2015flownet, mur2015orb, li2016vehicle, chen2017multi, yan2018second, lang2019pointpillars, yin2021centerpoint, liu2022bevfusion, zhou2023query, shi2022motion, dauner2023parting, yoo2026leveraging, huang2026feasibility}. More recently, the field has increasingly moved toward unified or end-to-end frameworks that jointly model multiple stages of the driving pipeline~\cite{hu2023uniad, jiang2023vad, weng2024paradrive, liao2025diffusiondrive, chen2025drivinggpt, liu2026drivepi, xia2026drivelaw, kirby2026drivor, li2026drivevla, zhang2025epona, zuo2026dvgt, zhang2026geoworldad}. This shift has expanded the ambition of autonomous driving research, but it has also increased the field's dependence on large, diverse, and well-integrated datasets for training, evaluation, and failure analysis.

The modern era of data-driven autonomous driving research was strongly shaped by KITTI~\cite{geiger2012kitti}, released in 2012, which provided benchmark tasks for driving perception using data collected in and around Karlsruhe, Germany. KITTI became a central reference point for stereo, optical flow, odometry, object detection, and related perception tasks~\cite{geiger2012kitti}, and is often regarded as a first generation of driving benchmark data~\cite{li2023open}. A second generation of autonomous driving datasets, led substantially by industry, expanded the scale, sensor modalities, annotation richness, and task coverage of benchmark data~\cite{caesar2020nuscenes, sun2020waymo, houston2021lyft, geyer2020a2d2}. For example, nuScenes provides a modern autonomous-vehicle sensor suite, including multi-view cameras, lidar, radar, GPS/IMU, and map information, and supports tasks beyond static perception, including tracking and trajectory prediction~\cite{caesar2020nuscenes}. Nevertheless, their spatial coverage remains limited to a small number of cities or regions.

Despite this limited coverage, these datasets have become de facto benchmarks for model development and evaluation. Even work explicitly concerned with domain adaptation, generalization, or new-city driving often remains anchored to the same standard releases~\cite{wang2020germany, yang2021st3d, xu2021spg, yuan2023bi3d, shin2025diffrefine, chen2024diffubox, lao2024lit, wang2026learning}. Similarly, recent efforts in vision-language understanding, safety-critical evaluation, and rare-scenario simulation often build new annotations or simulated tasks on top of canonical datasets, rather than incorporating a broader range of real-world data sources~\cite{sima2024drivelm, wang2025omnidrive, qian2024nuscenes, tian2025nuscenes, dauner2024navsim, cao2025pseudo, zhou2025hugsim, ljungbergh2024neuroncap, zeng2026rethinking}. This benchmark concentration is increasingly mismatched with the field's ambition: as driving models become more integrated and data-hungry, evaluation on a narrow set of canonical releases provides an incomplete view of geographic, operational, and long-tail generalization.

\section{The Long-Tailed Dataset Usage Phenomenon}
\label{sec:phenomenon}

Only recently have large industrial actors begun releasing datasets explicitly aimed at broader spatial and scenario coverage. Waymo's WOD-E2E curates challenging long-tail driving segments for end-to-end evaluation~\cite{xu2025wod}, while NVIDIA's Physical AI Autonomous Vehicles dataset provides large-scale multi-sensor driving data from thousands of cities across the US and Europe, spanning 25 countries~\cite{nvidia2025physicalai}. These efforts are valuable and encouraging, but they cannot by themselves solve the coverage problem. Even large industrial releases inevitably reflect the limits of a centralized data-collection paradigm: where companies choose to operate, which sensors they deploy, what data they can release, and which regions are commercially or strategically important to them. If the goal is autonomous driving that works \emph{anywhere, anytime, for anyone}, broad coverage cannot depend solely on a few centralized actors. This raises a natural question: can the research community complement industry by contributing distributed real-world driving datasets on a global scale?

\subsection{A Broad Community Data Effort with Long-Tailed Impact}
\label{sec:longtail_data}
To answer this question, we conducted a targeted literature review of datasets released with publications or technical reports, 
focusing on works that introduce new, independently collected real-world driving datasets. (See \autoref{sec:dataset_usage_analysis} for details.)
The resulting review yielded \textbf{647 papers} proposing real-world driving datasets between 2012 and 2026,
demonstrating that the research community has already sustained a decade-long, distributed effort to collect real-world driving data. The geographic coverage of these datasets has grown steadily, now spanning nearly 50 countries across all inhabited continents~(\autoref{fig:geo_combined}). This suggests that the community has both the capacity and the willingness to expand the data frontier beyond a small set of canonical benchmarks. 

\begin{figure}
    \centering
    \includegraphics[width=\linewidth]{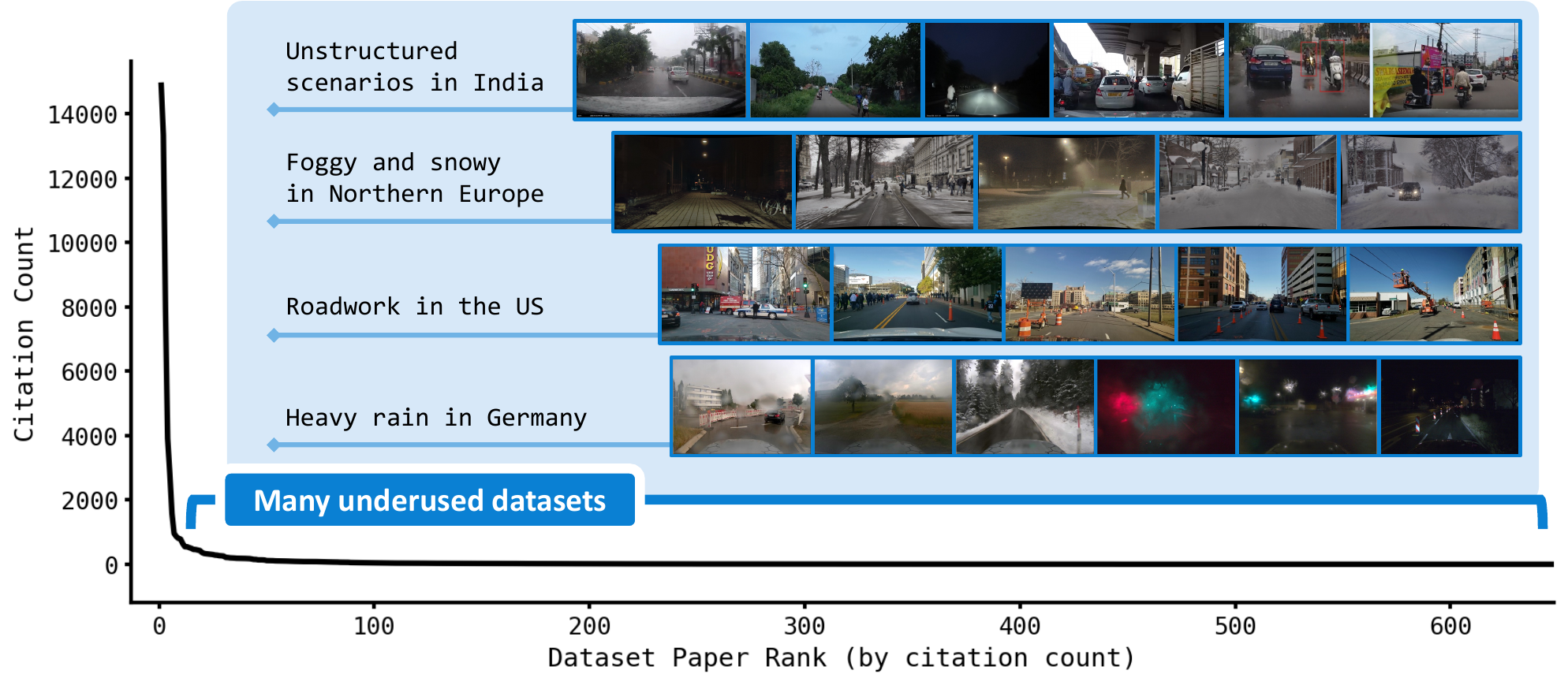}
    \vskip -5pt
    \caption{\textbf{Driving dataset usage is dramatically long-tailed.} Despite hundreds of available datasets, community adoption is concentrated on a few releases, indicating structural barriers to utilizing the broader collection---although underused datasets naturally provide diverse scenarios~\cite{wagner2026longtail,bijelic2020seeing, chandra2023meteor,ghosh2025roadwork}.} 
    \label{fig:longtail}
\end{figure}

Yet this broad collection effort has not translated into broad research impact. Dataset usage, measured by citation counts, follows a pronounced long-tailed distribution~(\autoref{fig:longtail}): a few releases account for the overwhelming majority of citations, while hundreds of datasets are cited rarely or not at all\footnote{Citations are an imperfect proxy for actual usage: some datasets may be used without being cited; others may be cited without being directly used. Still, they provide a useful signal of visibility and community influence.}. This pattern reveals a central tension: the community has produced a surprisingly broad set of real-world driving datasets, but the field has not effectively leveraged them. 

\subsection{The Underused Tail Contains Useful Signals}
\label{sec:diverse-data-useful}

The long-tailed phenomenon raises an important question: are the underused datasets genuinely useful, or are they peripheral because they add little beyond existing benchmarks? One possibility is that these datasets, despite their quantity and geographic spread, capture scenes that current models already handle well. After all, learning-based models are most challenged by out-of-distribution conditions; data from underexplored regions are not necessarily useful if their visual, behavioral, or operational distributions are already covered by mainstream benchmarks. We therefore conduct a preliminary study to probe whether datasets in the long tail can reveal meaningful model limitations.

\begin{figure}[t]
    \centering
    \includegraphics[width=\linewidth]{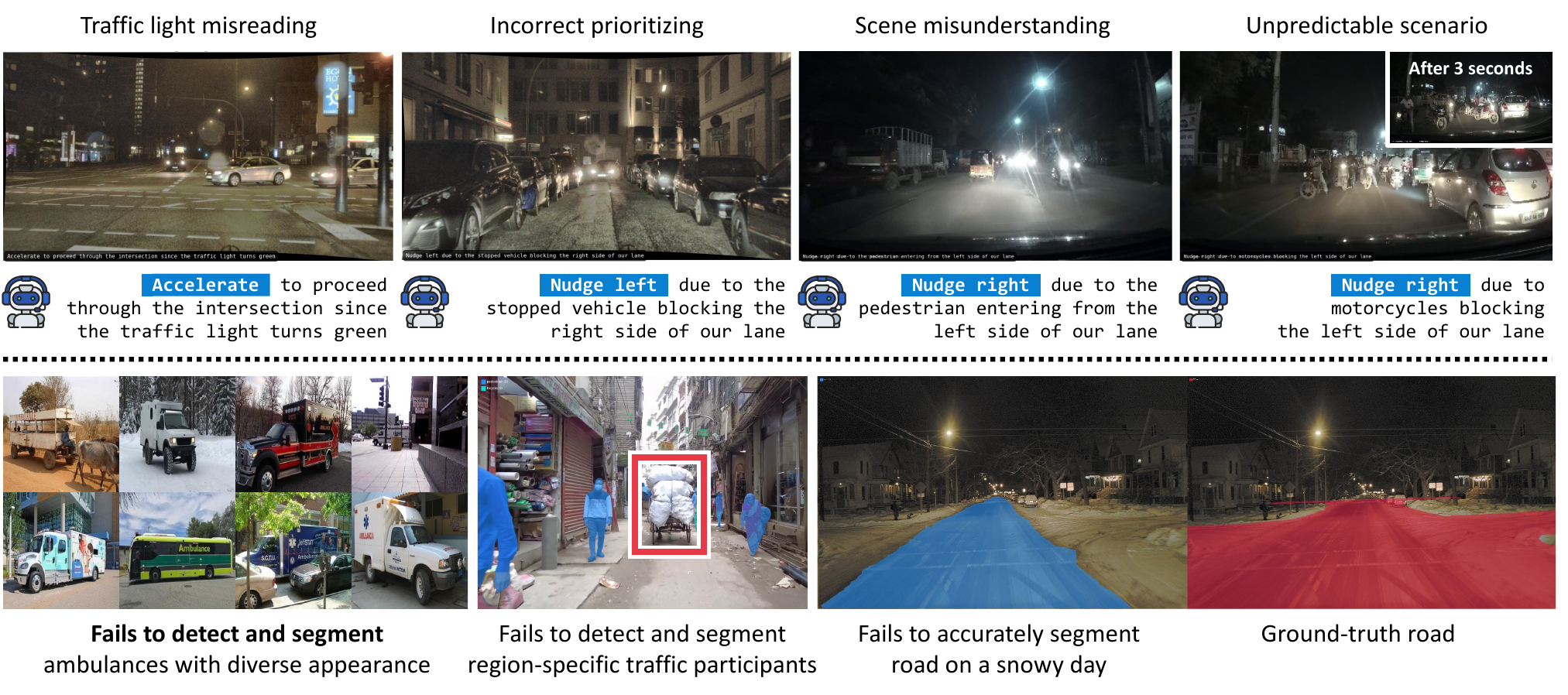}
    \vskip -5pt
    \caption{\textbf{State-of-the-art models fail systematically on available but underutilized datasets.}
        (\textit{Top}) Alpamayo 1.5~\cite{wang2025alpamayo}, evaluated on data from Northern Europe and India, exhibits reasoning failures.
        (\textit{Bottom}) SAM3~\cite{carion2025sam3}, evaluated on scenes featuring diverse appearances, fails to correctly segment objects that are routine in these settings but absent from standard benchmark distributions.}
    \label{fig:failure}
\end{figure}

\mypara{Geographic and Scenario Diversity Surface Model Failures.}
We first evaluate Alpamayo~1.5~\cite{wang2025alpamayo}, a recent end-to-end planning model, on two underutilized datasets without fine-tuning on either dataset: Seeing Through Fog~\cite{bijelic2020seeing}, collected in Northern Europe under adverse weather, and METEOR~\cite{chandra2023meteor}, collected in India. As shown in \autoref{fig:failure}~(\emph{Top}), the model exhibits systematic failures, including misreading traffic lights, incorrect agent prioritization, inaccurate scene understanding, and difficulty handling unpredictable interactions. We further evaluate SAM3~\cite{carion2025sam3}, a recent general-purpose segmentation model, on driving scenes containing objects, traffic participants, and road conditions underrepresented in standard benchmarks~\cite{gab2025innovative, diaz2022ithaca365,sahabangladeshi}. As shown in \autoref{fig:failure}~(\emph{Bottom}), the model fails to segment or detect several such elements, including regionally distinctive emergency vehicles and road surfaces under adverse weather. While qualitative, these examples indicate that underused datasets can expose model vulnerabilities not fully captured by canonical benchmarks.

\begin{wrapfigure}{r}{.4\linewidth}
  \centering
  \vskip-12pt
  \includegraphics[width=\linewidth]{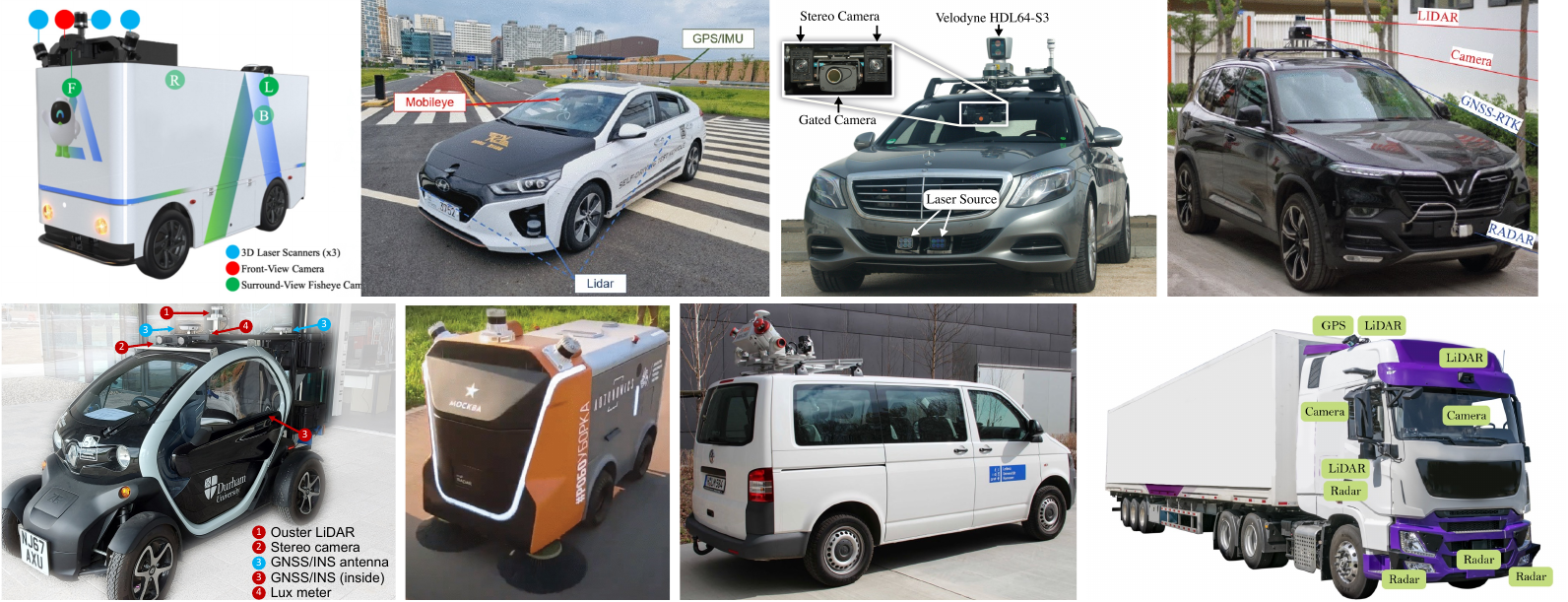}
  \vskip-5pt
  \caption{{Community-collected datasets are diverse in hardware configurations.}}
  \vskip-10pt
  \label{fig:vehicles}
\end{wrapfigure}

\mypara{Hardware Diversity Supports Cross-Platform Generalization.} Beyond scene diversity, community-contributed datasets also provide an underexplored source of hardware variation. Standard benchmarks are typically collected by vehicles sharing similar sensor suites and mounting configurations. Models trained and evaluated in this setting may therefore be sensitive to sensor changes; indeed, prior work shows that even small deviations in sensor placement can significantly degrade performance~\cite{kumar2025charm3r}. As shown in \autoref{fig:vehicles}, community-collected datasets span diverse sensor suites, mounting configurations, and vehicle types~\cite{loc2024accurate, wang2024esp, ni2025lane, julca2021gated3d, klokov2023daps3d, voelsen2021classification, li2021durlar, oh2022towards}. Incorporating such variation into training and evaluation could encourage models that generalize across hardware platforms rather than over-specializing to a single standardized setup. Such robustness would make models more deployable, reusable, and economically efficient in practice.

Together, these results suggest that datasets in the long tail are not merely numerous; they contain useful signals for evaluating and improving driving models. They expand spatial, scenario, environmental, object-level, and hardware diversity. Their limited use thus points to a deeper problem in the field's data ecosystem: useful data already exist, but the community lacks effective mechanisms to discover, integrate, evaluate on, and learn from them.
\section{Why Are They Underused? Limitations of the Existing Data Paradigm}
\label{sec:limitation}

\begin{figure}[t]
    \centering
    \includegraphics[width=\linewidth]{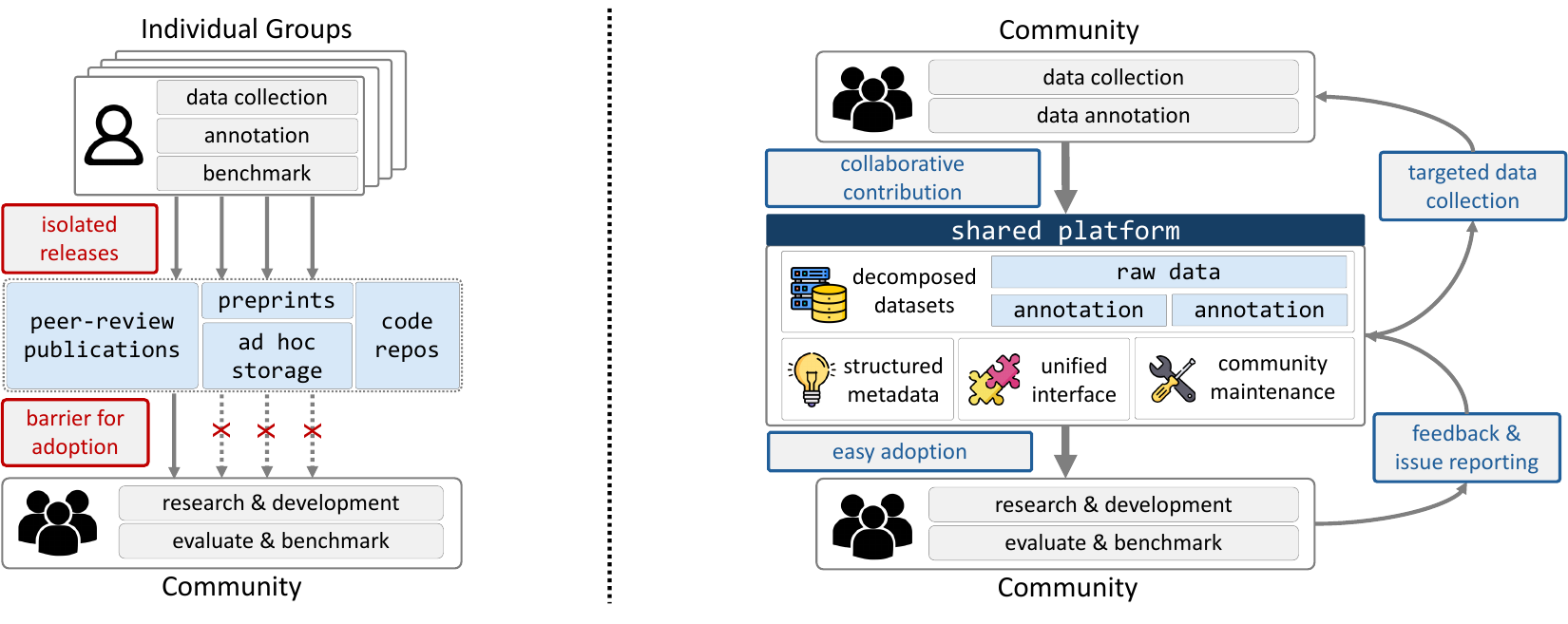}
    \vspace{-0.6em}
    \caption{Current~\vs~proposed community-driven data paradigm.}
    \label{fig:next_paradigm}
\end{figure}

We trace the structural barriers that prevent the field from leveraging its own distributed data efforts. The left side of~\autoref{fig:next_paradigm} illustrates the prevailing data paradigm: independent contributors produce their own datasets and are often responsible for the entire dataset lifecycle, from vehicle setup, data collection, and annotation to release, benchmarking, and maintenance. Yet only a small number of large, well-annotated, and highly visible benchmarks attract the community's attention, while the broader pool of available datasets remains underused. We next examine why this concentration persists; an expanded discussion is provided in \autoref{sec:challenge-paradigm-gap}.

\colorbox{blue!15}{\mypara{Challenge \#1: Finding Relevant Datasets Is Difficult.}}

The prevailing data paradigm often leads to isolated dataset releases. Even when suitable datasets exist, finding those that match a specific research need is far from straightforward. Low-profile datasets lack visibility, while even well-known ones scatter key information (\eg, modalities, annotations, licenses, and scenarios) across papers, repositories, and project pages. This creates a large, poorly organized search space that discourages adoption.

\colorbox{blue!15}{\mypara{Paradigm Gap: No Mechanism for Need-Based Dataset Discovery.}}

The current paradigm lacks structured, searchable dataset metadata and a unified protocol for recording key attributes. Researchers thus cannot easily query for specific needs, such as nighttime urban driving, rare weather, region-specific traffic participants, or particular sensor configurations. As a result, useful datasets are not surfaced to potential users, and the community lacks a systematic way to identify underrepresented scenarios or prioritize future data collection.

\colorbox{teal!15}{\mypara{Challenge \#2: Relevant Datasets May Have Limited Standalone Utility.}}

Even after a suitable dataset is identified, it may not be readily useful for the task at hand~\cite{liu2026openlongtail, pan2025typ}. Many datasets covering rare scenarios or underrepresented geographies are orders of magnitude smaller than dominant benchmarks such as Waymo Open Dataset~\cite{sun2020waymo} or nuScenes~\cite{caesar2020nuscenes}, limiting their use for robust training or evaluation. Others provide annotations that are too sparse or coarse for downstream tasks, such as trajectory prediction or fine-grained detection. Thus, valuable data may still offer limited practical benefit under existing workflows.

\colorbox{teal!15}{\mypara{Paradigm Gap: No Mechanism for Aggregating Contributions.}}

The current paradigm treats dataset creation as an isolated, end-to-end effort by each contributor. This makes large-scale collection expensive and fine-grained annotation difficult to sustain. Crucially, there is no mechanism for partial or collaborative contribution through which individually small efforts could accumulate into larger, more useful resources. As a result, many datasets remain structurally limited in scale, annotation richness, or task coverage.

\colorbox{violet!15}{\mypara{Challenge \#3: Poor Data Maintenance and Accessibility.}}

Independent maintenance also limits adoption. Even publicly released datasets may become inaccessible over time: download links break, hosting expires, documentation becomes outdated, and evaluation code stops running with current dependencies. Users therefore default to the small set of datasets that remain reliably hosted, documented, and actively maintained.

\colorbox{violet!15}{\mypara{Paradigm Gap: No Long-Term Support Structure for Datasets.}}

Long-term maintenance requires both infrastructure, such as durable storage and stable hosting, and labor, such as updating documentation, fixing loaders, and correcting annotation errors. Under the current paradigm, these responsibilities fall almost entirely on the original dataset creators, which is unrealistic as students graduate, researchers move on, and grants end. Without institutional or platform support, datasets are released as one-time artifacts rather than maintained as living resources.

\colorbox{orange!15}{\mypara{Challenge \#4: High Integration Cost Across Datasets.}}

Even when datasets are discoverable and accessible, incorporating them into an existing workflow requires substantial engineering effort~\cite{wang2020germany}. Driving datasets often differ in annotation schemas, coordinate frames, file formats, sensor conventions, and evaluation protocols, requiring custom loaders and converters before they can be used. This per-dataset integration cost discourages researchers from expanding beyond datasets that already fit their pipelines.

\colorbox{orange!15}{\mypara{Paradigm Gap: Isolated Releases Without Shared Infrastructure.}}

Each dataset is typically designed around its own sensors, annotations, and target tasks. While natural for individual releases, this sacrifices cross-dataset compatibility: standardized schemas, common abstractions, and stable evaluation interfaces are rarely treated as shared infrastructure. The ecosystem thus accumulates isolated datasets that are difficult to merge, substitute, or evaluate consistently.

Taken together, these challenges share a common root: the current paradigm often treats dataset release as an isolated, one-time effort. A group collects, annotates, and releases a dataset, but the ecosystem provides little support for discovery, aggregation, maintenance, or compatibility afterward. This explains why collective data collection has not yet translated into collective data utilization.

\section{Toward a Collaborative Community-Driven Data Paradigm}
\vskip-5pt

The aforementioned limitations call for a new paradigm: data and benchmark building should become a coordinated community effort, rather than a collection of isolated releases~(\autoref{fig:next_paradigm}, right). Community members could collaboratively contribute raw data, annotations, benchmarks, and evaluation protocols, while a persistent community-driven platform hosts these contributions with structured, searchable metadata and shared interfaces across datasets.

\subsection{Community-Driven Data Paradigm through Open and Shared Data Platform}
We elaborate on four design principles and their potential benefits below. Rather than prescribing a final design, our goal is to initiate discussion and help the field shape a data paradigm that is sustainable, extensible, and responsive to the community's evolving needs.

\colorbox{blue!15}{\mypara{Principle \#1: Structural Dataset Organization.}}

Because dataset information is currently fragmented across papers, project pages, and repositories, a collaborative data paradigm should make dataset organization explicit and structured. \textbf{\textit{Each contribution should therefore be accompanied by structured metadata}} (\eg, collection location, sensor configuration, weather conditions, scene type, supported tasks, annotation types, and licensing), exposed through a shared platform with sorting, filtering, and search capabilities.

This directly addresses dataset visibility: rather than manually parsing scattered papers, repositories, and project pages, users can query for the data they need. More importantly, shared metadata would give the community a systematic view of the data ecosystem itself: which scenarios are well covered, which remain underrepresented, and where current models repeatedly fail. These gaps and failures can then become signals for targeted contribution, making dataset growth responsive to model limitations and community needs rather than driven only by the priorities of individual groups.

\colorbox{teal!15}{\mypara{Principle \#2: Decomposed Data Collection and Annotation.}}

Under the current paradigm, a single group bears the full cost of data collection, annotation, and benchmark design, limiting both the scale and diversity of contributions. \textbf{\textit{We propose decoupling raw data and annotations as distinct, first-class contributions}}: one group can contribute raw data without annotating it, while another can annotate existing data without collecting its own.

This decomposition lowers the barrier to participation in two concrete ways. First, more groups can contribute data without bearing the full pipeline cost, broadening coverage across scenarios, regions, and conditions. Second, researchers can release annotations incrementally as new tasks emerge, allowing datasets to evolve with community needs. Together, these changes would allow individually modest contributions to accumulate into datasets that are larger in scale, richer in annotation, and more responsive to downstream research needs.

\colorbox{violet!15}{\mypara{Principle \#3: Dataset Maintenance as Shared Community Infrastructure.}}

Under the current paradigm, maintenance falls almost entirely on the original creators, causing datasets to decay as students graduate, projects end, and repositories go unmaintained. \textbf{\textit{A collaborative data paradigm should support long-term dataset maintenance through a persistent, community-driven platform}} that provides stable hosting, durable links, versioning, and documentation.

Beyond logistics, the platform should enable community-sourced upkeep, including issue reporting, annotation corrections, and patch submissions when tools break. 
Treating maintenance as shared infrastructure makes datasets more durable and valuable over time. Rather than disappearing when original creators move on, datasets can remain available, improve through continued contributions, and support new annotations, integrations, and tasks that did not exist at the time of collection.

\colorbox{orange!15}{\mypara{Principle \#4: Unified Standards Across Datasets.}}

Current driving datasets are released as disjoint resources, each with its own data formats, annotation schemas, and evaluation toolkits, with little consideration for cross-dataset compatibility. \textbf{\textit{A collaborative data paradigm should therefore promote unified standards across datasets to lower integration cost.}} Raw data, annotations, and evaluation interfaces should follow shared conventions per sensor type and task. These standards need not erase dataset-specific details, but should provide enough common ground to reduce per-dataset engineering overhead.

Lower integration cost directly encourages broader adoption. With common interfaces, researchers can more easily test models across diverse regions, conditions, and scenarios, allowing underrepresented datasets to enter mainstream evaluation. As their failure cases become visible, they can inform future model development. In this way, standardization and diversity reinforce each other.

\mypara{Remark.}
There already exist community-wise efforts like MMDetection3D~\cite{mmdet3d2020} and OpenPCDet~\cite{openpcdet2020}. In particular, recent efforts on 123D~\cite{dauner2026123d} provide a unified API for
heterogeneous driving data, representing an important step toward a community-driven open data platform.
These systems align with our Principle \#4 by reducing engineering cost through common data infrastructure, as well as Principle \#3 through active maintenance. Yet, they do not address issues such as data visibility and community contribution.
Our proposed paradigm can build on such foundations, connecting shared data interfaces with community-wide support for discovery, contribution, and long-term coordination.

\subsection{New Research Opportunities}\label{sec:new-research-opportunities}

Beyond addressing existing limitations, the proposed paradigm unlocks research directions that are currently difficult to pursue. 
The first direction is \emph{multi-sensor generalization}: building a single model that performs reliably across diverse sensor configurations is a practical yet largely unsolved question, with existing work limited to a few datasets or simulating sensor variation by artificially degrading inputs offline~\cite{wei2022lidar, peng2023cl3d, heylen2021monocinis, fang2024lidar}; a platform where each contributor brings their own vehicle and sensor setup provides a natural testbed on genuinely diverse hardware that no single dataset can offer. 
Next, \emph{offline labeling}---annotating raw data without real-time constraints---has attracted sustained industry interest~\cite{qi2021offboard, zhang2023oyster, caesar2021nuplan} and offers a distinct research direction in which models can leverage future frames, spend more compute per prediction, and exploit foundation models more aggressively; the continuously growing stream of raw contributor data provides a natural, ever-expanding testbed for this line of work. 
Finally, geographic expansion raises the question of how driving systems can keep learning from new environments~\cite{yoo2025rnbpop,xu2026civet}. \emph{Continual learning} is a realistic and practical challenge~\cite{jeon2026lessons, mai2022online, lesort2020continual} that remains underexplored in autonomous driving, partly due to a lack of suitable data~\cite{verwimp2023clad}; our paradigm aligns naturally with this setup, as periodic data releases can first serve as a held-out test set and then as additional training data, enabling a rolling evaluation protocol that mirrors real-world deployment. 

\subsection{Practical Issues}

Realizing the paradigm described above is not a technical problem that any single group can solve. It introduces many practical hurdles that require the community to reach an agreement on questions that extend well beyond engineering. We highlight four that demand early attention.

\mypara{Governance.}
A platform that loses funding or institutional support becomes inaccessible overnight. Who operates the platform, and how are disputes resolved? No single institution can hold unilateral control without creating conflicts of interest, yet fully decentralized governance is difficult to sustain.

\mypara{Licensing.}
Data contributed under incompatible license terms cannot be combined for evaluation or training, undermining the platform's core purpose. The community must agree on a licensing framework that balances contributors' rights with the need for broad downstream usability.

\mypara{Privacy.}
Driving data is particularly sensitive~\cite{zhang2025more}. Beyond faces and license plates, it can expose residential locations, personal travel routines, and pedestrian biometrics. Open platforms are especially prone to these risks, as contributors may not anticipate what their data reveals.

\mypara{Incentive.}
For a platform to sustain itself, contributors must receive meaningful credit. Under the current paradigm, recognition is mediated almost entirely through publication, leaving annotation, tooling, and maintenance invisible. It is essential to design community-driven recognition mechanisms, \eg, download counts, usage tracking, and community ratings.

\subsection{Exemplars from Other Communities}

The above challenges are not unique to autonomous driving, and other communities offer practical guidance. For instance, \textbf{iNaturalist}~\cite{van2018inaturalist, mckinley2017citizen} provides the closest model: a citizen science platform that has sustained millions of biological observations from contributors worldwide. It addresses governance through a multi-stakeholder steering committee, licensing through a structured default (CC BY-NC) with a set of options. Also, \textbf{HuggingFace} and \textbf{GitHub} demonstrate that open contribution platforms can scale through low barriers to entry and strong visibility mechanisms, such as download counts, community ratings, and peer review. Both suggest that visible credit for contributors is a powerful driver of sustained participation.
Lastly, \textbf{The Event Horizon Telescope (EHT)}~\cite{event2019first} demonstrates what globally distributed scientific collaboration can achieve. The project was carried out by nearly 140 institutions across 19 countries, linking many existing radio observatories to form a single Earth-sized virtual telescope. The key enabler was community-wide agreement on data formats, synchronization protocols, and processing pipelines across institutions with very different resources and governance structures. We believe this is a direct precedent for what the autonomous driving community could achieve with equivalent coordination.

\section{Alternative Views}
\label{sec:alternative}

The most direct counterargument to our position is that a new data paradigm is unnecessary and that universal autonomous driving can be achieved within the current paradigm, especially as industry continues to release larger-scale and more challenging datasets. We do not dismiss this view entirely; an industry-led paradigm can plausibly lead the field toward its goals, \emph{but only if industry has the incentive to do so} across the full diversity of conditions that universal autonomous driving requires.
We argue that this incentive is structurally limited. For reasons of safety, regulatory compliance, and profitability, industry deployments are naturally concentrated in regions and scenarios where the business case is strongest. A company expanding its robotaxi service has little commercial motivation to collect data from rural Southeast Asia or adverse-weather conditions in sub-Saharan Africa. The result is a paradigm that, even at its best, optimizes for the distribution that matters to industry, not the full distribution that the field's stated goals demand.
This is precisely where the research community, and academia in particular, can provide unique and complementary value. Academic groups operate under different incentives: they are motivated by scientific coverage, generalization, and the kinds of challenging edge cases that industry may deprioritize. A community-driven paradigm does not replace industry releases---it addresses the parts of the problem that industry cannot. We include an expanded discussion in~\autoref{sec:appendix_alternative}.

\section{Conclusion, and a Call for Participation}
\label{sec:conclusion-call}
Autonomous driving research does not merely need larger datasets; it needs a community-governed data ecosystem. Existing datasets already cover a surprisingly broad range of locations and scenarios, yet their impact is bottlenecked by fragmentation, incompatible protocols, and benchmark incentives that concentrate attention on a few dominant datasets. Progress toward anytime-anywhere autonomy requires treating data collection, failure discovery, annotation, and benchmarking as shared community infrastructure.
To be clear, the proposed paradigm in this position piece is not intended to replace large-scale industry releases, but to \emph{complement} them. The community can leverage industry-collected data for training while using diverse community-collected data to evaluate generalization and surface failures. Industry benefits from richer signals in safety-critical scenarios; community contributors gain credit and visibility. The two paradigms reinforce each other.
The autonomous driving community has spent over a decade building better algorithms. It is time to also invest in the data paradigm those models depend on. There is no shortage of venues to debate architectures and leaderboards, but no established forum where the community seriously discusses how driving data should be collected, maintained, and shared. We call on industry, academia, annotation companies, and governments to create that forum, not to compete, but to agree on the foundations that will determine whether the field's ambitions are realized. The models will be ready to generalize. The question is whether the community is ready to give them the data to do so.

\mypara{Acknowledgments.}
This research is supported in part by grants from the National
Science Foundation (IIS-2107077, IIS-2107161, CNS-2211599, IIS-2305532).
We thank the Ohio Supercomputer Center for the generous support
of the computational resources.


\bibliographystyle{splncs04}
\bibliography{main}

\newpage

\appendix

\section{Appendix}
\label{sec:supplementary}

\subsection{Discussion With and Against Alternative Views}
\label{sec:appendix_alternative}

\mypara{The Problem May Be Standardization Rather Than a New Paradigm.}
A second view is that the community may not need a fundamentally new data paradigm. Perhaps the main problem is simply that existing datasets need better documentation, more consistent metadata, more reliable hosting, and standardized formats. If dataset authors followed stronger release practices, researchers might naturally make better use of the long tail of available data.

We view standardization as a necessary component of the proposed paradigm. However, standards are difficult to sustain when every dataset is released independently. Each group must make separate decisions about file organization, annotation format, metadata, licenses, baseline code, and maintenance. A shared platform can reduce this burden by providing common templates, unified interfaces, and searchable metadata. In this sense, the platform is not a substitute for standards; it is a mechanism for making standards practical, visible, and reusable across the community.

\mypara{The Long Tail May Contain Datasets of Uneven Quality.}
Another critical view is that the long-tailed usage pattern may be rational. If researchers continue to use a few dominant datasets, it may not be because the community lacks synergy, but
because those datasets are simply more reliable. From this view, underuse is not a failure of the
data paradigm; it is a quality filter.

Currently, the community has no systematic way to determine which underused datasets are low quality, which are specialized but valuable, and which could reveal important model failures if integrated into evaluation. A community-driven paradigm should not equate availability with usefulness. Instead, it should make quality and scope more transparent.  Community feedback, ratings, download counts, and usage in published work can provide additional signals, but they should be combined with technical validation and clear documentation. The goal is not to promote every dataset, but to distinguish genuinely valuable long-tail data from unusable data in a transparent and
community-legible way.

\mypara{Community Incentives May Be Hard to Design.}
Another alternative view is that a community platform may not solve the incentive problem. Data collection and annotation are expensive, and researchers may still prioritize algorithm papers over data contributions. Ratings and downloads may not be enough to reward the labor of collecting, cleaning, documenting, and maintaining datasets. Without meaningful credit, contributors may not participate.

A successful paradigm must treat raw data, annotations, metadata,  maintenance, evaluation protocols, and failure reports as key contributions. The platform should make these contributions visible and citable, and research venues should consider how to recognize them. This is one reason community-wide discussion is necessary: the platform alone cannot change incentives unless conferences, journals, institutions, and industry partners also value data work. The goal is to create an ecosystem in which good contributions accumulate credit rather than disappear after a one-time dataset release.

\subsection{Details on Dataset Usage Analysis}
\label{sec:dataset_usage_analysis}

\begin{wrapfigure}[15]{r}{0.51\textwidth}
    \vspace{-2.em}
    \centering
    \includegraphics[width=0.51\textwidth]{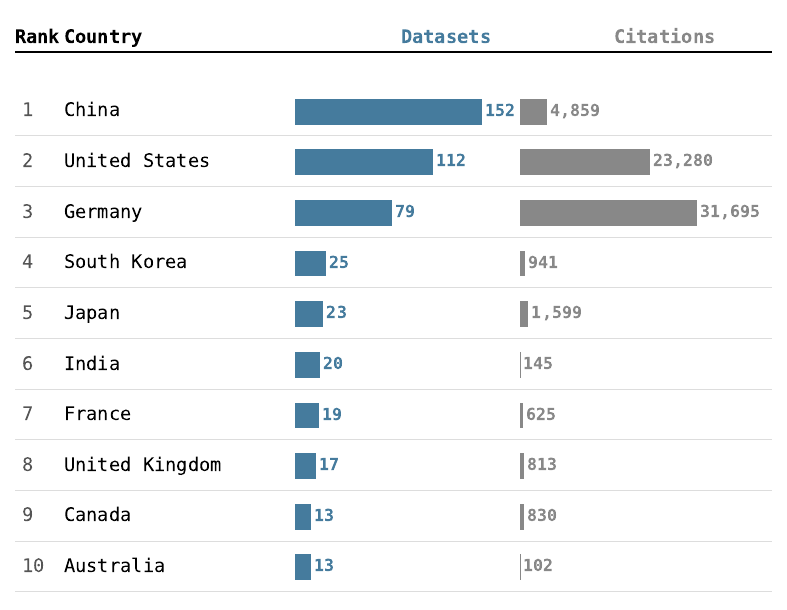}
    \vspace{-15pt}
    \caption{Top 10 countries by datasets and citations.}
    \label{fig:country_table}
\end{wrapfigure}
To analyze the dataset usage, we leverage dataset publication and corresponding citation count as a surrogate metric. Here, we detail our targeted literature review methodology.
Specifically, we queried Semantic Scholar for papers citing five highly cited driving-scene datasets---KITTI~\cite{geiger2012kitti}, Cityscapes~\cite{cordts2016cityscapes}, nuScenes~\cite{caesar2020nuscenes}, the Waymo Open Dataset~\cite{sun2020waymo}, and BDD100K~\cite{yu2020bdd100k}. 
These datasets serve as broad anchors for the post-2012 driving-data literature: while not exhaustive, we expect that many subsequent dataset papers would cite at least one of them. This search retrieved over 26,000 citing papers. We then applied a two-stage filtering pipeline: first, an LLM-based classifier identified papers that appeared to introduce a genuinely new, independently collected real-world driving dataset; second, manual verification removed derivative works, simulation-only datasets, and re-annotations of existing data. This process yielded 642 papers proposing real-world driving datasets between 2012 and 2026, which we refer to in \autoref{sec:longtail_data} of the main text.

\mypara{Dataset Paper Classification Prompt.}
To identify dataset papers from candidate papers retrieved via citation search, we used the Claude Batch API (\texttt{claude-haiku-4-5}). We classified each paper by its title and abstract. Below are the exact system and user prompts used.

\begin{tcolorbox}[colframe=gray!50!black, colback=gray!10!white, title=System Prompt, breakable]
\small

You are a research paper classifier specializing in autonomous driving and computer vision.

\medskip

Your task: given a paper title and abstract, determine whether the paper introduces a \textbf{genuinely new, independently collected real-world driving dataset} as its primary or major contribution.

\medskip

\textbf{YES} --- classify as a new driving dataset paper if:

\begin{itemize}[leftmargin=1.5em]
    \item The paper collects brand-new sensor data from real vehicles driving on public roads (cameras, lidar, radar, GPS/IMU, or combinations thereof)
    \item The data is freshly captured by the authors --- not derived, re-labeled, or reformatted from existing datasets
    \item The dataset targets any autonomous driving or road-scene perception task: object detection, semantic/instance/panoptic segmentation, depth estimation, optical flow, scene flow, tracking, lane detection, localization, trajectory prediction, pedestrian behavior, adverse weather perception, vehicle-to-vehicle/infrastructure perception, etc.
    \item It is acceptable if the paper also proposes a method or benchmark alongside the new dataset
\end{itemize}

\textbf{NO} --- do NOT classify as a new driving dataset paper if:

\begin{itemize}[leftmargin=1.5em]
    \item The paper uses existing datasets (KITTI, nuScenes, Waymo, Cityscapes, BDD100K, Argoverse, Lyft, A2D2, ONCE, Apolloscape, etc.) without contributing new collected data
    \item The paper re-annotates, re-labels, augments, or post-processes an existing dataset
    \item The paper creates a subset, filtered split, or curated version of an existing dataset
    \item The paper merges or concatenates multiple existing datasets
    \item The paper introduces a synthetic-only dataset (simulation, game engines, neural rendering, diffusion-generated data)
    \item The paper's primary contribution is a method, model, or algorithm --- even if it includes minor data collection
    \item The dataset domain is not road/driving scenes (medical imaging, drones-only, indoor, satellites, industrial)
    \item The paper is a survey, benchmark evaluation, or challenge report without new data collection
\end{itemize}

\textbf{Borderline guidance:}

\begin{itemize}[leftmargin=1.5em]
    \item If a paper collects even a small new real-world driving dataset as a major contribution alongside a method, answer YES
    \item Mixed real+synthetic: answer YES only if the real component is substantial and independently valuable
    \item Datasets from vehicles on private/test tracks still count if they capture real sensor data
\end{itemize}

Respond in valid JSON only. No text outside the JSON.

\end{tcolorbox}

\begin{tcolorbox}[colframe=gray!50!black, colback=gray!10!white, title=User Prompt (per paper), breakable]
\small

\texttt{Title: \{title\}}

\medskip

\texttt{Abstract: \{abstract\}}

\medskip

Classify this paper. Respond with ONLY valid JSON:

\begin{verbatim}
{
  "is_driving_dataset_paper": true or false,
  "is_real_world": true or false,
  "dataset_name": "name of dataset if proposed, else null",
  "confidence": "high" or "medium" or "low",
  "reason": "one sentence explaining your decision"
}
\end{verbatim}

\end{tcolorbox}

\mypara{Additional Analysis.} \autoref{fig:country_table} summarizes the geographical distribution of dataset contributions and their accumulated citations among the 647 papers. The United States and Germany dominate in terms of citation impact, reflecting the outsized influence of a small number of landmark datasets originating from these countries. China is the leading contributor by sheer volume of datasets, but the comparatively lower citation counts for Chinese datasets may partly reflect differences in data accessibility and licensing constraints that limit their adoption by the broader international community. More broadly, the distribution reveals a pronounced long-tail phenomenon in dataset usage: while a handful of datasets from a few well-resourced countries accumulate the vast majority of citations, a large number of datasets covering diverse geographical regions and driving conditions remain severely underutilized, disproportionate to their potential value for building robust, globally generalizable perception systems.

\subsection{VLM-Based Semantic Tagging on Underused Datasets}
\label{app:vlm_annotation}

Vision-language models (VLMs) have the potential to assist semantic
annotation of driving data, such as the tagging of weather and
other scene attributes for data organization and retrieval.
We examine this application as another setting in which underused
driving datasets can provide useful test cases for identifying
limitations of current models.

Specifically, we conduct a weather-tagging evaluation on Seeing
Through Fog~\cite{bijelic2020seeing}, a dataset collected across
northern Europe with frame-level weather annotations.
We sample 100 frames for each combination of four weather conditions
(snow, dense fog, light fog, and clear) and two times of day
(day and night), yielding 800 frames in total. 
We evaluate Qwen3-VL-2B-Instruct~\cite{qwen3technicalreport}, Qwen3-VL-8B-Instruct~\cite{qwen3technicalreport}, and
Cosmos-Reason2-8B~\cite{nvidia2025cosmosreason28b} using a prompt that requests JSON fields for
weather, time of day, road surface, visibility, and confidence.
We use generic weather categories rather than requiring the models
to reproduce dataset-specific labels: predictions of \texttt{fog}
are counted as correct for both \texttt{light\_fog} and
\texttt{dense\_fog}. 
We observe two insights from the results elaborated below.

\begin{table}[h]
    \centering
    \small
    \renewcommand{\arraystretch}{1.05}

    \begin{minipage}[t]{0.33\textwidth}
        \vspace{0pt}
        \centering

        \begin{minipage}[t][6\baselineskip][t]{\linewidth}
            \caption{Weather-label accuracy (\%) of
            Qwen3-VL-8B-Instruct.
            Each cell contains 100 frames.}
            \label{tab:vlm_weather}
        \end{minipage}
        \vskip-15pt
        \begin{tabular}{lrr}
            \toprule
            Condition & Day & Night \\
            \midrule
            Snow      & 89.0 & 90.0 \\
            Dense fog & 96.0 & 65.0 \\
            Light fog & 56.0 &  9.0 \\
            Clear     & 75.0 & 21.0 \\
            \bottomrule
        \end{tabular}
    \end{minipage}
    \hfill
    \begin{minipage}[t]{0.64\textwidth}
        \vspace{0pt}
        \centering
        \setlength{\tabcolsep}{3pt}

        \begin{minipage}[t][6\baselineskip][t]{\linewidth}
            \caption{Weather-label accuracy (\%) on light and
            dense fog across three models.
            Both conditions use \texttt{fog} as the target label.
            Each cell contains 100 frames.}
            \label{tab:vlm_fog}
        \end{minipage}
        \vskip-15pt
        \begin{tabular}{lcccc}
            \toprule
            & \multicolumn{2}{c}{Day}
            & \multicolumn{2}{c}{Night} \\
            \cmidrule(lr){2-3}
            \cmidrule(lr){4-5}
            Model & Light & Dense & Light & Dense \\
            \midrule
            Qwen3-VL-8B-Instruct & 56.0 & 96.0 & 9.0 & 65.0 \\
            Cosmos-Reason2-8B    & 46.0 & 96.0 & 4.0 & 51.0 \\
            Qwen3-VL-2B-Instruct & 45.0 & 95.0 & 5.0 & 46.0 \\
            \bottomrule
        \end{tabular}
    \end{minipage}
\end{table}

\paragraph{Failures under specific weather conditions.}
Table~\ref{tab:vlm_weather} shows substantial variation across
conditions. For Qwen3-VL-8B-Instruct, accuracy on light fog drops
from 56\% during the day to 9\% at night.
Table~\ref{tab:vlm_fog} shows that this difficulty appears across
all three evaluated models: nighttime light-fog accuracy is
9\%, 4\%, and 5\% for Qwen3-VL-8B-Instruct,
Cosmos-Reason2-8B, and Qwen3-VL-2B-Instruct, respectively.
Since both light and dense fog are mapped to the same target label,
these results reveal failures to recognize fog under light-fog
conditions, even without requiring distinctions in fog severity.

\paragraph{Potential mismatches in label interpretation.}
The results also suggest that some disagreements may involve
how weather categories are interpreted. Of the 200 frames annotated
as clear, Qwen3-VL-8B-Instruct predicts \texttt{snow} for 53.
In 47 of those 53 cases, the model also predicts a
\texttt{snow\_covered} road surface. This pattern suggests that
some disagreements may reflect differing interpretations of
falling snow and snow already present on the ground.

This study illustrates how underused driving data
can also be useful for evaluating VLM-based semantic tagging.
The evaluated dataset exposes difficulties in recognizing
specific weather conditions and potential mismatches between
model predictions and dataset annotation conventions.
As VLMs are explored for driving-data annotation and retrieval,
underused datasets can provide valuable cases for examining
their capabilities and limitations.

\subsection{Challenges and Limitations of Current Data Paradigm}
\label{sec:challenge-paradigm-gap}
Core to our position is that the current data paradigm is insufficient to address the needs of the research community, thus leading to the underutilization symptom. We include an extensive discussion of the challenges and the corresponding paradigm gap below.

\colorbox{blue!15}{\mypara{Challenge \#1: Finding Relevant Datasets is Difficult.}}

Today, one of the observed pain points in using datasets is simply finding relevant ones. Some datasets are less visible because they are not created by high-profile groups or released through high-visibility venues such as top conferences, challenges, or competitions. Others are visible, yet their key details---modality, domain, annotation type, license, scale, or evaluation protocol---are fragmented across papers, repositories, project pages, and dataset cards. While suitable datasets may exist and be publicly available, locating them within a vast and poorly organized search space is a challenge in its own right.
Consider a researcher studying autonomous driving in adverse weather conditions. They may be unaware that relevant datasets exist outside of the well-known benchmarks, simply because those datasets were released by smaller research groups without a dedicated project page or conference spotlight. Even if they do find a candidate dataset, determining whether it contains the right sensor configuration, annotation type, or geographic coverage requires manually cross-referencing multiple sources. In practice, most researchers default to familiar datasets not because they are the best fit, but because they are the easiest to find and evaluate.
This difficulty is not incidental---it reflects a structural gap in the current paradigm. Dataset details are rarely exposed as structured, searchable metadata, leaving no effective way to query for specific needs (\textit{e.g.}, nighttime urban driving, rare weather conditions, or a particular sensor configuration). Beyond individual discovery, the community also lacks a systematic way to identify which scenarios are underrepresented or which data gaps should be prioritized for future collection.

\colorbox{teal!15}{\mypara{Challenge \#2: Relevant Datasets May Have Limited Standalone Utility.}}

Small-scale, research-effort datasets today can still fall short for the task at hand, even if they are identified.
A first common limitation is scale. Datasets covering rare scenarios or underrepresented geographies are often orders of magnitude smaller than predominant ones~\cite{sun2020waymo,caesar2020nuscenes}, making them useful for small-scale studies but unsuitable for robust training or benchmarking.
A second limitation is annotation mismatch. Even when a dataset captures a valuable domain, its annotations may not align with the researcher's needs. Consider an Indian driving dataset that captures dense, heterogeneous urban traffic---a scenario of genuine research interest. If it only provides detection annotations but lacks ground truth for surrounding agent trajectories, it cannot be used for planning research. Worse, if the raw sensor data is not released alongside the annotations, there is no way to retroactively generate the missing labels. Similarly, a dataset may have the right annotation type but at too coarse a temporal resolution---\textit{e.g.}, keyframes annotated at 2Hz when the task requires dense per-frame labels.
These limitations are not coincidental---they reflect a structural barrier to large-scale contribution in the current paradigm. Dataset creation is treated as an isolated, end-to-end effort in which a single group bears the full cost of collection and annotation. This coupling makes large-scale datasets prohibitively expensive to produce, and fine-grained annotation often infeasible even after data is collected. Critically, there is no mechanism for partial or collaborative contribution, through which individually small efforts could be pooled into a larger, more useful resource. As a result, datasets are structurally constrained in scale and annotation granularity from the outset.

\colorbox{violet!15}{\mypara{Challenge \#3: Poor Data Maintenance and Accessibility.}}

Dataset maintenance is frequently a challenge; even when datasets are publicly released, they often become inaccessible over time. Download links break, hosting expires, documentation grows outdated, and evaluation code stops running with current dependencies. This is especially common for datasets released by students or non-professional contributors, who may no longer be actively maintaining the project after graduation or a change in research focus.
The root cause is structural. Long-term maintenance has two components: logistical (\textit{i.e.}, durable storage, stable hosting, persistent download links) and labor (\textit{i.e.}, updating documentation, fixing broken loaders, correcting annotation errors). Under the current paradigm, both fall entirely on the original dataset creators---an unrealistic expectation given that students graduate, researchers move on, and grants end. There is no persistent platform or institution to absorb these responsibilities, so datasets are effectively released as one-time artifacts. The consequence is that the community converges on a small set of well-resourced datasets that happen to have the institutional backing to remain accessible, rather than those that are most relevant.

\colorbox{orange!15}{\mypara{Challenge \#4: High Integration Cost Across Datasets.}}

One other barrier to democratizing data collection is that incorporating them into an existing workflow requires substantial per-dataset engineering effort. Driving datasets commonly differ in annotation schemas, coordinate frames, file formats, and evaluation protocols, requiring custom loaders and converters before any data can be used. For example, a researcher wishing to evaluate a detection model across three geographically diverse datasets may find that each defines its own 3D bounding box convention, uses a different lidar coordinate frame, and reports metrics with a different IoU threshold---meaning the integration cost must be paid separately for each dataset before any meaningful comparison is possible.
This reflects a broader absence of shared release conventions in the current paradigm. Each dataset is designed around what is most natural for its own sensors and target tasks, which is convenient for individual releases but sacrifices cross-dataset compatibility. The ecosystem thus accumulates isolated datasets that are difficult to merge, substitute, or evaluate consistently---and the practical cost of each additional dataset discourages users from expanding beyond the few that already fit their pipelines.

\end{document}